\documentclass{bmvc2k}
\pdfoutput=1

\usepackage{wrapfig,lipsum,booktabs}
\usepackage{amsmath}
\usepackage{amssymb}
\usepackage{mathtools}
\usepackage{amsthm}
\usepackage{graphicx}
\usepackage{enumitem}
\usepackage{xcolor, colortbl}
\usepackage{arydshln}

\usepackage{hyperref}
\hypersetup{
    colorlinks=true,
    linkcolor=blue,
    filecolor=magenta,      
    urlcolor=cyan,
    pdftitle={Overleaf Example},
    pdfpagemode=FullScreen,
    }

\newcommand{\tksmall}[1]{{\tiny$\pm$ #1}}

\newcommand{\ing}[1]{\textcolor{blue}{\tiny ($\uparrow$ #1)}}
\newcommand{\decg}[1]{\textcolor{blue}{\tiny ($\downarrow$  #1)}}
\newcommand{\ind}[1]{\textcolor{orange}{\tiny ($\uparrow$ #1)}}
\newcommand{\decd}[1]{\textcolor{orange}{\tiny ($\downarrow$ #1)}}

\newcommand\Tstrut{\rule{0pt}{2.2ex}}         
\newcommand\Bstrut{\rule[-0.9ex]{0pt}{0pt}}   
\usepackage[capitalize,noabbrev]{cleveref}

\definecolor{Gray}{gray}{0.90}
\definecolor{LightCyan}{rgb}{0.88,1,1}

\title{Unifying Synergies between Self-supervised Learning and Dynamic Computation}

\addauthor{Tarun Krishna}{tarun.krishna2@mail.dcu.ie}{1,3}
\addauthor{Ayush K. Rai}{ayush.rai3@mail.dcu.ie}{1,2}
\addauthor{Alexandru Drimbarean}{Alexandru.Drimbarean@xperi.com}{4}
\addauthor{Eric Arazo}{eric.arazo@insight-centre.org}{1,3,5}
\addauthor{Paul Albert}{paul.albert@insight-centre.org}{1,3,5}
\addauthor{Alan F. Smeaton}{alan.smeaton@dcu.ie}{1,2}
\addauthor{Kevin McGuinness}{kevin.mcguinness@dcu.ie}{1,3}
\addauthor{Noel E. O'Connor}{noel.oconnor@dcu.ie}{1,3}

\addinstitution{
 Insight Centre for Data Analytics,\\
 Dublin City University,\\
 Dublin, Ireland
}
\addinstitution{
 School of Computing,\\
 Dublin City University,\\
 Dublin, Ireland
}
\addinstitution{
 School of Electronic Engineering,\\
 Dublin City University,\\
 Dublin, Ireland
}
\addinstitution{
 Xperi Corporation,\\
 Galway, Ireland\\
}
\addinstitution{
 Equal Contribution
}

\runninghead{Krishna \emph{et al}}{Unifying SSL and Dynamic Computation}
\begin{document}

\maketitle

\begin{abstract}

\noindent Computationally expensive training strategies make self-supervised learning (SSL) impractical for resource-constrained industrial settings. 
Techniques like \textit{knowledge distillation} (KD), \textit{dynamic computation} (DC), and \textit{pruning} are often used to obtain a lightweight model, which usually involves multiple epochs of fine-tuning (or distilling steps) of a large pre-trained model, making it more computationally challenging.
In this work we present a novel perspective on the interplay between the SSL and DC paradigms.
In particular, we show that it is feasible to simultaneously learn a \textit{dense} and \textit{gated} \underline{sub-network} from scratch in an SSL setting without any additional fine-tuning or pruning steps. 
The co-evolution during pre-training of both dense and gated encoder offers a good accuracy-efficiency trade-off and therefore yields a generic and multi-purpose architecture for application-specific industrial settings.
Extensive experiments on several image classification benchmarks including CIFAR-10/100, STL-10 and ImageNet-100, demonstrate that the proposed training strategy provides a \textit{dense} and corresponding \textit{gated} sub-network that achieves on-par performance compared with the vanilla self-supervised setting, but at a significant reduction in computation in terms of FLOPs, under a range of target budgets ($\text{t}_{d}$).
\end{abstract}

\section{Introduction}
\label{sec:intro}
\textbf{Motivation.} Self-supervised representation learning methods \cite{Chen2020, Caronb, Chenb, Chenf, Bardes} are the standard approach for training large scale deep neural networks (DNNs). One of the main reasons for their popularity is their capability to leverage the inherent structure of data from a vast unlabeled corpus during pre-training, which makes them highly suitable for transfer learning \cite{goyal2021self}. 
However, this comes at the cost of substantially larger model size, computationally expensive training strategies (larger training times, large batch-sizes, etc.) \cite{chen2020big, goyal2021self} and subsequently more expensive inference times. 
Though such strategies are effective for achieving state-of-the-art results in computer vision, they may not be practical in resource-constrained industrial settings that require lightweight models to be deployed on edge devices.

  To lessen the computational burden, it is common to extract (or learn) a \textit{lightweight} network from an off-the-shelf pre-trained model. This has been successfully achieved through techniques such as knowledge distillation (KD) \cite{hinton2015distilling}, pruning \cite{frankle2018lottery}, dynamic computation (DC) \cite{veit2018convolutional}, etc. KD methods follow a standard two-step procedure of pre-training and distilling knowledge into a \textit{student} network using self-supervised (SS) objective \cite{Fang,abbasi2020compress, navaneet2021simreg} or by together incorporating supervised and SS objectives \cite{Tian}, while pruning based approaches heavily rely on multiple steps of  pre-train$~\rightarrow$~prune~$\leftrightarrow$~finetune to get a lightweight network irrespective of the objective, whereas methods based on dynamic/conditional computation \cite{veit2018convolutional, herrmann2020channel} again rely on a pre-trained model to obtain a \textit{lightweight} network while keeping the network topology intact via a \textit{gating} mechanism. These approaches are effective but using fine-tuning to obtain a sub-network from large pre-trained models (such as Large Language Models) can be computationally expensive and cumbersome. Also, since downstream tasks are diverse and vary widely, any change in the task requires repeating the entire procedure multiple times, making it inefficient and less transferable. 

\noindent \textbf{Research Questions.} These limitations motivate us to ask the following question: \textit{``Can we unify the learning of a lightweight sub-network along with a dense network from scratch and in a completely self-supervised fashion?''} A straightforward way to achieve this is via an \textit{online} KD (with self-supervised objective) \cite{yang2023online, bhat2021distill} learning paradigm which involves training teacher ($f_{\theta}$) and student ($g_{\phi}$) networks simultaneously during a pre-training stage. Recognising that this adds to the computational burden during pre-training (extra $g_{\phi}$), we adopt a different route to attain the same goal but with a simpler, efficient pre-training objective and faster inference than online KD-based methods. 
\begin{figure*}[t]
  \centering
    \includegraphics[width=1.00\linewidth]{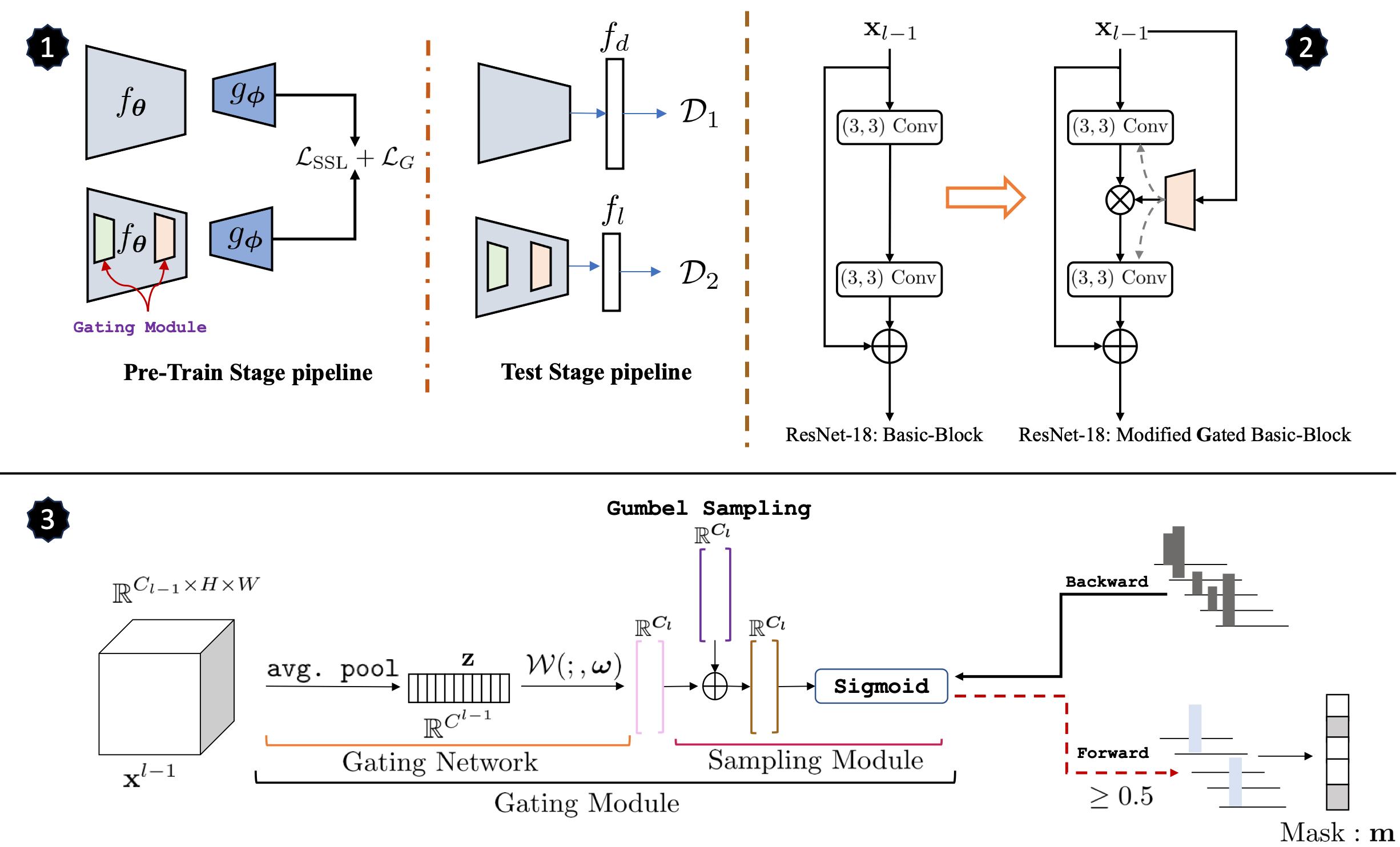}
    \vspace{-1.75\baselineskip}
    \caption{\textbf{1.} Illustrates the unification of SSL and DC during the pre-training and testing (inference) phase. $f_{d}$ and $f_{l}$ denotes respective linear layer for dense and gated network respectively. \textbf{Note:} dimensional size of $f_{d}$ and $f_{l}$ is the same, while in the figure this may look otherwise but is done to depict the fact that empirically, the dimension of $f_{l}$ is less than $f_{d}$. \textbf{2.} Shows the modification of ResNet-18 basic block to accommodate the gating network during inference. \textbf{3.} Describes the gating module which comprises a \textit{gating} network and \textit{sampling} module.}
    \label{fig:main} 
    
\end{figure*}

This enables us to reformulate the research question to: ``\textit{Can we learn a single encoder (function) that could serve the dual purpose of being used as a dense and lightweight network with minimal additional overhead?}''

Our objective is to simultaneously learn a \textit{dense} and a \textit{lightweight} model through a unified pre-training procedure to maintain high performance on the downstream task. We achieve this by  exploiting the Siamese setting (a common setting for SSL \cite{balestriero2023cookbook}) combined with a gating mechanism for dynamic channel selection  (DCS) \cite{veit2018convolutional,Li_2021_ICCV}.  We opted for dynamic channel selection over KD/pruning for two main reasons: first, the gating mechanism preserves the networks topology adding enough flexibility to the approach; second, these gating modules are computationally inexpensive. For the self-supervised objective we choose VICReg \cite{Bardes} due to its symmetric nature and its ability to regularise each branch independently as dense and sparse branches will have different statistics.  Figure~\ref{fig:main} (\textit{left}) demonstrates this dual setting of obtaining a dense and a lightweight network (derived from the dense one).

It should be noted that in this paper we do not follow the vocabulary of student-teacher networks. Instead we restrict the terminology to lightweight (gated)\footnote{we use lightweight and gated networks interchangeably.} and a dense network, where encoder and gates are randomly initialised and trained jointly from scratch, with the aim that they co-evolve during pre-training. It is, however, important to mention that in this work we are not proposing any new KD objective or KD-induced learning algorithm nor any new DC objective or any pruning-based learning. Instead we provide a novel perspective on exploiting the synergies that exist between self-supervised representation learning and dynamic computing. This approach is easily extendable to other symmetric-twins like Barlow-Twins \cite{Zbontar2021}, SimCLR \cite{Chen2020} or W-MSE \cite{Zbontar2021}, while it may require some adjustments for non-symmetric methods like BYOL \cite{grill2020bootstrap}, MoCo \cite{he2020momentum}, etc., which will be explored in  future work. Our main contributions are:

\begin{itemize}[]
\item We present a novel perspective of unifying the learning of \noindent \textit{dense} and \textit{lightweight} networks by exploiting a symmetric joint embedding architecture of the SSL paradigm.
\item We demonstrate that a single encoder can be exploited as a dense as well as a lightweight network; we show in Table~\ref{table1} and Table~\ref{table2} that a single base encoder can serve this \underline{dual} purpose. This not only reduces computational overhead during training but also gives enough flexibility to use a single network and exploit it as per its requirement.
\item We demonstrate exhaustively, through experiments that this unification preserves feature quality across different experimental settings and gives on-par performance when compared with strict baselines (Section~\ref{result}). 
\end{itemize}

\section{Background and Related Work}

\textbf{Self-supervised representation learning}:
SSL methods can be broadly divided into contrastive and non-contrastive techniques in the current scenario.  At the core of these approaches is the concept of learning joint embedding representations realised via a Siamese \cite{1467314} architecture through instance discrimination. Contrastive learning (CL) \cite{wu2018unsupervised, oord2018representation, Henaff2020, Chen2020, he2020momentum, tian2020makes,9226466} based approaches intuitively try to bring similar instances closer while contrasting them with negative samples. Non-contrastive techniques include clustering methods \cite{caron2018deep,asano2019self, Caronb}, which alternate between cluster assignment and predicting clusters as pseudo labels.
Approaches such as BYOL \cite{grill2020bootstrap}, OBoW \cite{gidaris2020online}, MoCo \cite{he2020momentum}, and SimSiam \cite{chen2021exploring} use a \textit{teacher}-\textit{student} approach to learn joint representations.
While approaches like Barlow Twins \cite{Zbontar2021}, W-MSE \cite{Ermolov}, and VICReg \cite{Bardes} follow a more principled approach of information maximisation.  We use VICReg for SSL because it can regulate each branch independently, which is more effective when each branch has different statistics.

\noindent \textbf{Dynamic Computation}: 
DC is a resource-efficient mechanism that reduces model complexity by skipping unimportant parts of the network while preserving the networks topology. Several authors including \cite{figurnov2017spatially,mcgill2017deciding,huang2018multi,wu2020dynamic,wang2018skipnet} have proposed adding decision branches to different layers of convolutional neural networks (CNN) for learning early exiting strategies leading to faster inference. BlockDrop \cite{Wu_2018_CVPR} and SpotTune \cite{guo2019spottune} learn a policy network to adaptively route the inference path through fine-tuned or pre-trained layers. 
ConvNet-AIG \cite{veit2018convolutional}, \cite{herrmann2020channel} introduced a network that adaptively selects specific layers of importance to execute depending on the input image by specifying a target rate of each layer. GaterNet \cite{chen2019you} introduced a network to generate input-dependent binary gates to select filters in the backbone network. DGNet \cite{Li_2021_ICCV} proposed a dual gating mechanism to induce sparsity along spatial and channel dimensions. Furthermore, dynamic channel pruning methods have also been devised such as feature boosting and suppression (FBS) \cite{gao2019dynamic} to dynamically amplify and suppress output channels computed by CNN layers. Other works learn sparsity through a three-stage pipeline: \textit{pretrain-prune-finetune} as in \cite{tiwari2020chipnet} or use pre-trained models. See \cite{hoefler2021sparsity} for a more detailed explanation of sparsity, pruning and dynamic computing.\\ 
\noindent \textbf{Self-supervised dynamic computation and beyond}:
Most of the works on dynamic computation have been confined to supervised learning. Recently, \cite{krishna2022dynamic} used SimSiam \cite{Chenf} as a self-supervised objective combined with a dynamic channel gating (DGNet) \cite{Li_2021_ICCV} mechanism 
and showed that comparable performance can be achieved under channel budget constraints. 
Likewise \cite{9878423} used a channel gating-based dynamic pruning (CGNet) \cite{hua2019channel} augmented with CL to achieve inference speed-ups without substantial loss of performance. 
 In a similar line of work, \cite{chen2021lottery} used iterative magnitude pruning (IMP) to obtain a winning ticket \cite{frankle2018lottery} for a pre-trained task (self-supervised objective) and evaluated its performance on various downstream tasks. \cite{9856963} extended the work done in \cite{caron2020pruning} in a MoCo (pre-trained) setting
augmented with ADMM \cite{zhang2018systematic} for systematic pruning.
A self-supervised \textit{loss} objective can serve as a tool for KD \cite{hinton2015distilling} and model compression (MC). \cite{tian2019contrastive} used a contrastive objective (along with a supervised loss for \textit{task specific distillation}) to train a student network from a pre-trained network. Similar to \cite{tian2019contrastive} but in a completely self-supervised setting  \cite{abbasi2020compress, fang2021seed} minimises the $KL$-divergence between the distribution of similarities for the teacher (pre-trained) and student networks, while SimReg \cite{navaneet2021simreg} minimises the regression loss. The authors in \cite{xu2020knowledge} used a two-step strategy to train a teacher (with labels and then using an SSL head with a fixed backbone) followed by training a student using a KD loss. However, we follow a more simplistic approach through the unification of SSL (VICReg) and DCS, where DCS maintains the network topology, making fine-tuning easier on different downstream tasks, unlike other methods that make network structure irreversible.
\section{Preliminaries and Setup} \label{prelim}
\textbf{1. VICReg as SSL objective:} VICReg \cite{Bardes} learns a joint embedding space governed by a loss objective, which consists of \textit{invariance} ($s$) (mean squared error (MSE)), \textit{variance} ($v$) and \textit{co-variance} ($c$), depicted Equation \ref{eq1}. Let us consider some image dataset $\mathbf{X} = \{\mathbf{x}_{i}\}_{i=1}^{\mathcal{D}}$
and a set of transformations $\mathcal{T}$ (refer to supplementary material for details). An anchor image $\mathbf{x}_{i} \in \mathbb{R}^{H \times W \times 3}$ is 
 augmented through transformations $t_{1}, t_{2} \sim \mathcal{T}$ to get $\mathbf{x}^{1}_{i} = t_{1}(\mathbf{x}_{i})$ and $\mathbf{x}^{2}_{i} = t_{2}(\mathbf{x}_{i})$ respectively. Augmented views are encoded through $f_{\theta}$ (ResNet-18 \cite{he2016deep} (R18) in this study) to get feature representations. Furthermore, these representations are mapped to an \textit{embedding} space via \textit{expander} ($g_{\phi}$) where the final VICReg loss is applied between the embedding vectors $\mathbf{z}_{i}^{1} = g_{\phi}(f_{\theta}(\mathbf{x}^{1}_{i}))$ and $\mathbf{z}_{i}^{2} = g_{\phi}(f_{\theta}(\mathbf{x}^{2}_{i}))$. Formally the loss is defined on a batch of embedding vectors $\mathbf{Z^{1}} = [\mathbf{z}_{1}^{1},\cdots,\mathbf{z}_{|\mathcal{B}|}^{1}]$ and $\mathbf{Z^{2}} = [\mathbf{z}_{1}^{2},\cdots,\mathbf{z}_{|\mathcal{B}|}^{2}]$ as:  
\begin{equation}\label{eq1}
    \begin{aligned}
        \mathcal{L}_{\text{VICReg}}(\mathbf{Z}^{1}, \mathbf{Z}^{2}) &={} \overbracket{\underbracket{\mu [v(\mathbf{Z}^{1})+ v(\mathbf{Z}^{2})]}_{\text{\textbf{V}ariance}}
    + \underbracket{\nu[c(\mathbf{Z}^{1}) + c(\mathbf{Z}^{2})]}_{\text{\textbf{C}o-Variance}}}^{\text{Regularisation Term}} + \underbracket{\eta s(\mathbf{Z}^{1}, \mathbf{Z}^{2})}_{\text{\textbf{I}nvariance}}, 
    \end{aligned}
\end{equation}
where $\mu = 25$, $\nu =25$ and $\eta = 1.0$. For detailed description of Equation \ref{eq1} refer to \cite{Bardes}.

\noindent \textbf{2. Gating for channel selection.} 
The gating module comprises of a \textit{gating network} \cite{Li_2021_ICCV, bejnordibatch} and a \textit{sampling module} \cite{jang2016categorical} ( Figure \ref{fig:main}). The \textit{gating network} can be thought of as a lightweight  network that decides the \textit{relevance} of channels referred to as \textit{importance} vector. 
To enable a lightweight design of the gating network ($\mathcal{W}$) we follow the squeeze and excitation block design \cite{hu2018squeeze}, similar to \cite{veit2018convolutional, herrmann2020channel}. This usually requires obtaining a context vector $\mathbf{z} \in \mathbb{R}^{C_{l-1}}$ via global average pooling to accumulate spatial information. This context vector $\mathbf{z}$ is processed through $\mathcal{W}$ to get relevance scores for each channel:

\begin{equation}
  \mathcal{W}(\mathbf{z}, \boldsymbol{\omega}) = \mathbf{w}_{2}*(\texttt{BatchNorm}(\textbf{w}_{1} * \textbf{z}))_{\text{ReLU}}, \quad \{\textbf{w}_{1}, \textbf{w}_{2}\} \in \omega 
\end{equation}

\noindent where $*$ denotes convolution, $\mathbf{w}_{1} \in \mathbb{R}^{\frac{C_{l-1}}{r} \times C_{l} \times 1 \times 1 }, \mathbf{w}_{2} \in \mathbb{R}^{C_{l} \times \frac{C_{l-1}}{r} \times 1 \times 1}$ and $r$ is defined as reduction rate (set to 4). 

Finally, to make a selection over a subset of a channels, we need to map the output of $\mathcal{W}$ to a binary vector (or mask $\mathbf{m} \in \mathbb{R}^{C_{l}}$). This discrete selection works perfectly during inference but breaks the computational graph during training. To make training possible, the \textit{sampling module} utilises the Gumbel-softmax reparameterisation trick \cite{jang2016categorical} to make this discrete selection without breaking the computational graph. Figure \ref{fig:main} (\textit{right 2}) shows the modification of ResNet18 \textit{basic block} (during inference).

In this work we follow the setting of DGNet \cite{Li_2021_ICCV} for channel selection  where sparsity is induced by setting a global target budget ($t_{d}$) to optimise a loss objective : 
 \begin{equation}\label{sparloss}
     \mathcal{L}_{G} = \lambda  \Bigg(\frac{\sum_{l=1}^{L}F^{R}_{l}}{\sum_{l=1}^{L}F_{l}^{O}} -t_{d}\Bigg)^{2} 
 \end{equation}
 where $F_{l}^{R}$ is the average FLOPs over the batch along with FLOPs contribution from the gating network $\mathcal{W}$ (which is fixed for each layer), while $F_{l}^{O}$ is the original FLOPs without a gating module, $\lambda =5$ \cite{Li_2021_ICCV} across all  datasets and training regimes. Only blocks with gating modules take part in FLOP computation as they contribute to the sparsification of the network. 
 We refer our approach as VICReg-Dual-Gating (\textbf{VDG}).


\subsection{Experimental Setup and Implementation Details }

\textbf{1. Pre-training:} We closely follow the implementation of VICReg \cite{Bardes} (as our self-supervised objective) suited with our computational constraints using solo-learn library \cite{da2022solo}, while for dynamic gating we follow DGNet \cite{Li_2021_ICCV} for inducing channel sparsity via gating mechanism with ResNet18. The unified framework is depicted in Figure~\ref{fig:main}: the two branches have a \textit{separate batch normalisation} layer following \cite{yu2018slimmable}. The encoder  and gating networks are \textit{randomly} initialised and trained with SGD for 500 epochs with a batch size of 512 on 2 Nvidia 2080Ti GPUs, with a warmup start of 10 epochs following a cosine decay with a base learning rate of 0.3 using the LARS optimizer \cite{you2017large}. Since we are using a very lightweight model as our gating network, there is no significant computational overhead during training, to be precise there is slight increase in computation which amounts to 2.11\% of extra model parameters (0.013\% of FLOPs computation). We pre-train for a target budget, $t_{d} = \{10\%, 30\%, 50\% \}$ for each of the datasets except for ImageNet-100 where $t_{d}$ is restricted to $\{30\%, 50\% \}$ due to computational constraints. We report the inference speedup in terms of a hardware-independent theoretical metric of FLOPs and not wall-clock time as we do not avail any hardware accelerators to utilise sparsity during training. 
Any scaling parameter in the loss term is derived from the respective paper. The code is available at \url{https://github.com/KrishnaTarun/Unification}.

\noindent \textbf{2. Evaluation:} Pre-training and evaluation is carried on the train and validation data of CIFAR-10/100 \cite{krizhevsky2009learning}, ImageNet100 \footnote{ \url{https://www.kaggle.com/datasets/ambityga/imagenet100}} and STL-10 \cite{coates2011analysis}. For pre-training with STL-10 we considered only the \textit{un-labelled} set. 
We follow the standard practice of evaluating the trained encoder by freezing its weights and training a linear classifier on top of it. We trained a single linear layer for 100 epochs with a batch size of 512 on a single NVIDIA 2080Ti with a learning rate of 0.3 following step decay of 0.1 at 60$^{\text{th}}$ and 80$^{\text{th}}$ epochs. We report top-1 accuracy averaged over 5 runs. 


\noindent \textbf{3. Baselines}: To exhaustively compare the performance of the dense and gated models we consider VICReg \cite{Bardes} as a SSL \underline{\textit{dense}} baseline  while VICReg augmented with sparsity loss $\mathcal{L}_{G}$ (following Krishna \textit{et al.}~\cite{krishna2022dynamic}) serves as a \underline{\textit{gated}} baseline, here goal is to train a lightweight gated encoder from scratch with a self-supervised objective.

\noindent \textbf{4. Comparison with self-supervised KD:} In order to make a fair assessment of this unification, we compare the gated network's performance with KD based methods specifically SEED \cite{fang2021seed} and SimReg \cite{navaneet2021simreg} where the former was proposed to distill representational knowledge into a smaller network  while the latter showed that a simple regression objective can serve as an effective tool for knowledge transfer. For SimReg: During pre-training (SSL) (500 epoch) we used VICReg \cite{Bardes} with a ResNet-18 (R18: Teacher) as base encoder trained on CIFAR-100 and ImageNet-100 ($\mathcal{D_{\text{Pretrain}}}$), while distillation is performed using the SimReg objective on $\mathcal{D}_{\text{Target}}$ ($=\mathcal{D_{\text{Pretrain}}}$) by further training it for another 130 epochs. For SEED: We use a MoCo-v2\footnote{\url{https://github.com/facebookresearch/moco}} pre-trained ResNet50 (R50) encoder trained on ImageNet-1K ($\mathcal{D_{\text{Pretrain}}}$), while distillation using the SEED objective is performed for 200 epochs on $\mathcal{D}_{\text{Target}}$ (same as SimReg). Student networks are derived by sampling from R18's subspace with the number of filters for each \textit{basic block} derived from our gating network i.e, channels are selected following policy learned by our gating module (see Figure (2,3) and Section 2 in supplementary) for fair comparison. The representation from the last average pooling layer is $l_{2}$ normalised and is evaluated  using kNN as the evaluation criterion with $k = 1$ to report top-1 accuracy. 
\section{Results}\label{result}

\textbf{1. Quantitative assessment.} Table~\ref{table1} compares the performance of VDG with the other two baselines for dense and gated. The lightweight gated network achieves improved performance across all datasets and target budgets ($t_{d}$) as compared to \textit{Baseline-2}, with a negligible drop at $t_{d}=50\%$ for CIFAR-10 only.
The improved gated performance can be attributed to the fact that both dense and lightweight gated models co-learn during pre-training via weight sharing. However, the performance gain is compensated by a slightly smaller reduction in FLOPs as compared to \textit{Baseline-2}. Improved performance of the gated network further closes the gap with a completely self-supervised dense model (\textit{Baseline-1}). 
In comparison to \textit{Baseline-1}, we observe minor drop in performance of VDG e.g., at $t_{d}=10\%$ across CIFAR-10 \decd{2.17\%}, STL-10 \decd{2.86\%}, CIFAR-100 \decd{1.55\%}, ImageNet-100 \decd{2.7\%} but with a significant reduction in the number of FLOPs. This illustrates that even under severe budget constraints our model achieves comparable performance to \textit{Baseline1}. This drop further decreases on increasing $t_{d}$ to $50\%$.

\begin{table*}[t]
 \caption{Linear Evaluation: $\uparrow$/$\downarrow$  in \textcolor{orange}{orange} font is comparison with \textit{Baseline-1}, while \textcolor{blue}{blue} font is comparison with \textit{Baseline-2}. FLOPs R. denotes FLOP reduction. We report Top-1 accuracy averaged over 5 runs. Best viewed in color. }
 \small
 \centering
 \resizebox{1.00\linewidth}{!}{
 \begin{tabular}{l|cc|c|cc|ccc}
      
      \hline
      \hline
      \multicolumn{1}{l|}{\textbf{}} & \multicolumn{2}{c|}{\textbf{VICReg} } & \multicolumn{1}{l|}{} &\multicolumn{2}{c|}{\textbf{VICReg-Gating}}&
      \multicolumn{3}{c}{\textbf{VICReg-Dual-Gating}} \Tstrut\Bstrut\\

     \multicolumn{1}{l|}{\textbf{}} & \multicolumn{2}{c|}{ \textcolor{orange}{Baseline-1}} & \multicolumn{1}{l|}{} &\multicolumn{2}{c|}{\textcolor{blue}{Baseline-2}}&
      \multicolumn{3}{c}{\textit{this work}} \Tstrut\Bstrut\\ 
\multicolumn{1}{l|}{\textbf{}} & \multicolumn{2}{c|}{Bardes \textit{et al.}~\cite{Bardes}} & \multicolumn{1}{l|}{} &\multicolumn{2}{c|}{Krishna \textit{et al.}~\cite{krishna2022dynamic}}&
      \multicolumn{3}{c}{} \\
      \textbf{Dataset}& Dense &FLOPs &$\textbf{t}_{d}$(\%) & \multicolumn{1}{c}{Gated} &\multicolumn{1}{c|}{FLOPs R.} &\multicolumn{1}{c}{Dense $\uparrow$} &\multicolumn{1}{c}{Gated $\uparrow$ } &FLOPs R. $\uparrow$ \Bstrut \\
\hline
&&&10\%& 87.75 \tksmall{0.03} & 85.92\% &88.99 \tksmall{0.04} \decd{2.12} & 88.94 \tksmall{0.06} \decd{2.17} \ing{1.19} & 81.49\% \decg{4.43}  \Tstrut\Bstrut\\
CIFAR-10&91.11 \tksmall{0.03}  &7.03E8 & 30\% & 89.49 \tksmall{0.04} & 69.27\% &90.38 \tksmall{0.04} \decd{0.73}  & 90.27 \tksmall{0.03} \decd{0.84} \ing{0.78} & 66.43\% \decg{2.84} \\
&&&50\%& 90.70 \tksmall{0.04} & 51.62\% & 90.20 \tksmall{0.02} \decd{0.91} & 90.40 \tksmall{0.06} \decd{0.71} \decg{0.30} & 49.02\% \decg{2.60} \Bstrut\\
\hline

&&&10\%& 82.48 \tksmall{0.15} & 82.85\% & 84.29 \tksmall{0.21} \decd{1.86} &83.29 \tksmall{0.05}  \decd{2.86} \ing{0.81} & 78.34\% \decg{4.51}\Tstrut\Bstrut\\
STL-10&86.15 {\tiny$\pm$ 0.10} &3.33E8 & 30\% &84.16 \tksmall{0.11} & 68.38\%& 84.90 \tksmall {0.05} \decd{1.25}  &84.85 \tksmall{0.04} \decd{1.30} \ing{0.69} & 65.24\% \decg{3.14}\\
&&&50\%& 85.40 \tksmall{0.20} & 49.93\% & 85.75 \tksmall{0.02} \decd{0.40} & 85.72 \tksmall{0.02} \decd{0.43} \ing{0.32} & 48.41\% \decg{1.52} \Bstrut\\
\hline

&&&10\%& 63.12 \tksmall{0.09} & 84.82\% &65.21 \tksmall{0.06} \decd{0.65} &64.31 \tksmall{0.08} \decd{1.55} \ing{1.19} & 81.71\% \decg{3.11} \Tstrut\Bstrut \\
CIFAR-100&65.86 {\tiny$\pm$ 0.10} &7.03E8 & 30\% &65.41 \tksmall{0.09}& 68.68\%&65.90 \tksmall{0.10} \ind{0.04} &65.64 \tksmall{0.00} \decd{0.22} \ing{0.23} & 66.83\% \decg{1.85} \\
&&&50\%& 65.75 \tksmall{0.12} & 50.04\% &66.41 \tksmall{0.05} \ind{0.55} & 66.40 \tksmall{0.14} \ind{0.54} \ing{0.65} &49.06\% \decg{0.98} \Bstrut\\
\hline

&&& 30\% &74.04 \tksmall{0.09}& 67.95\%&75.12 \tksmall{0.07} \decd{2.62} &75.04 \tksmall{0.10} \decd{2.17} \ing{1.00} & 64.98\% \decg{2.97}\Tstrut\Bstrut\\
ImageNet-100& 77.74 {\tiny$\pm$ 0.12} & 1.81E9&50\%& 75.83 \tksmall{0.07}& 50.11\% & 76.42 \tksmall{0.26} \decd{1.32} &76.24 \tksmall{0.12} \decd{1.51} \ing{0.41} & 47.69\% \decg{2.42} \Bstrut\\
 \hline
 \hline
 \end{tabular}
 \label{table1}
 }
 \end{table*}

\begin{table*}[t]
 \caption{Transfer Performance: dense and gated under VICReg-Dual-Gating is compared with the common dense \textit{baseline} of VICReg. $\uparrow/\downarrow$ represents increment/decrement in performance. We report Top-1 linear evaluation accuracy averaged over 5 runs.}
 \small
 \centering
 \resizebox{1.00\linewidth}{!}{
 \begin{tabular}{ll|c|cc|cc|cc}
      \hline
      \multicolumn{2}{l|}{\textbf{Dataset}}  &\multicolumn{1}{c|}{\textbf{VICReg}} & \multicolumn{6}{c}{\textbf{VICReg-Dual-Gating}}
       \Tstrut\Bstrut\\
       \cline{4-9}
\multicolumn{2}{l|}{\textbf{}} & \multicolumn{1}{l|}{Bardes \textit{et al.}~\cite{Bardes}} &\multicolumn{2}{c|}{10\%}&\multicolumn{2}{c|}{30\%}&\multicolumn{2}{c}{50\%} \Tstrut\\
      \multicolumn{2}{l|}{From $\rightarrow$ To} &Dense  &\multicolumn{1}{c}{Dense}& \multicolumn{1}{c|}{Gated} & \multicolumn{1}{c}{Dense}& \multicolumn{1}{c|}{Gated} &\multicolumn{1}{c}{Dense}& \multicolumn{1}{c}{Gated} \Bstrut \\
\hline
CIFAR-100 $\rightarrow$ STL-10&& 70.37&64.69 $\downarrow$&64.29 $\downarrow$ &65.63 $\downarrow$ &65.93 $\downarrow$ &66.52 $\downarrow$ &67.30 $\downarrow$   \Tstrut\\
CIFAR-100 $\rightarrow$ CIFAR-10&& 80.08&80.56 $\boldsymbol{\uparrow}$&79.92 $\downarrow$&80.23 $\boldsymbol{\uparrow}$ &80.24 $\boldsymbol{\uparrow}$ & 80.16 $\boldsymbol{\uparrow}$ &80.09 $\boldsymbol{\uparrow}$ \\
CIFAR-100 $\rightarrow$  ImageNet-100&& 39.45 & 35.88 $\downarrow$ & 36.31 $\downarrow$ & 38.49 $\downarrow$ & 38.20 $\downarrow$ & 39.41 $\downarrow$ & 40.68 $\boldsymbol{\uparrow}$ \Bstrut\\
\hline
 ImageNet-100 $\rightarrow$ STL-10 &&72.80&-&-&65.86 $\downarrow$ &65.01 $\downarrow$  &67.74 $\downarrow$ &67.15 $\downarrow$   \Tstrut\Bstrut\\
ImageNet-100 $\rightarrow$ CIFAR-10&&55.12&-&-& 53.38 $\downarrow$ &49.41 $\downarrow$ &54.01 $\downarrow$ &49.80 $\downarrow$ \\
ImageNet-100 $\rightarrow$ CIFAR-100&& 31.42 & -& -&28.81 $\downarrow$ &25.55 $\downarrow$ &28.75 $\downarrow$ &25.14 $\downarrow$ \Bstrut\\

\hline
\end{tabular}
\label{table2}
}
\end{table*}
%

\begin{table*}
 \caption{Comparison of KD methods \textit{students} performance with our \textit{gated} network.}
 \small
 \centering
 \resizebox{1.00\linewidth}{!}{
 \begin{tabular}{l|lccc|c }
      \hline
      \multicolumn{1}{l|}{\textbf{Method}}  &\multicolumn{1}{c}{$\mathcal{D}_{\text{Pretrain}}$} & \multicolumn{1}{c}{$\mathbf{\mathcal{D}}_{\text{Target}}$} & \multicolumn{1}{c}{\text{SSL-Pre Method $(\text{Teacher})_{\text{epoch}}$}} &   \multicolumn{1}{c|}{\text{Student}} &  \multicolumn{1}{c}{\text{1-NN}} 
       \Tstrut\Bstrut\\      
\hline
 & CIFAR-100 & CIFAR-100 & VICReg $(\text{R}18)_{500}$ & R18 (10\%) & 37.32\%\\
  & CIFAR-100 & CIFAR-100 & VICReg $(\text{R}18)_{500}$ & R18 (30\%)& 45.71\%\\
 SimReg \cite{navaneet2021simreg}& CIFAR-100 & CIFAR-100 & VICReg $(\text{R}18)_{500}$ & R18 (50\%)&48.99\%\\ 
 \cdashline{2-6}
 & ImageNet-100 & ImageNet-100 & VICReg $(\text{R}18)_{500}$ & R18 (30\%)&63.80\%\\
 & ImageNet-100 & ImageNet-100 & VICReg $(\text{R}18)_{500}$ & R18 (50\%)&65.78\%\\ 
\hline
& ImageNet-1K & CIFAR-100 & MoCO-v2 $(\text{R}50)_{800}$ & R18 (10\%)&31.54\%\\
  & ImageNet-1K & CIFAR-100 & MoCO-v2 $(\text{R}50)_{800}$ & R18 (30\%)&35.29\%\\
 SEED \cite{fang2021seed}&ImageNet-1K & CIFAR-100 & MoCO-v2 $(\text{R}50)_{800}$ & R18 (50\%)&38.22\%\\ 
 \cdashline{2-6}
 & ImageNet-1K & ImageNet-100 & MoCO-v2 $(\text{R}50)_{800}$ & R18 (30\%)&64.38\%\\
 & ImageNet-1K & ImageNet-100 & MoCO-v2 $(\text{R}50)_{800}$ & R18 (50\%)&67.50\%\\ 
 \hline
 & CIFAR-100 & - & VICReg $(\text{R}18)_{500}$ & R18 (10\%) & \textbf{56.57\%}
\\
  {}Ours&  CIFAR-100 &- & VICReg $(\text{R}18)_{500}$ & R18 (30\%) & \textbf{58.49\%}\\
 (Gated)& CIFAR-100 & - & VICReg $(\text{R}18)_{500}$ & R18 (50\%)&\textbf{59.83\%}\\ 
 \cdashline{2-6}
 & ImageNet-100 & - & VICReg $(\text{R}18)_{500}$ & R18 (30\%) & \textbf{65.54\%}\\
 & ImageNet-100 & - & VICReg $(\text{R}18)_{500}$ & R18 (50\%) & \textbf{67.72\%}\\
 \hline
\end{tabular}
\label{tableKD}
}
\end{table*}
Another important aspect of our learning method is the performance of the \textit{dense} ($f_{\theta}$) model. Ideally our aim is to achieve fewer fluctuations with varying $t_{d}$ with a performance equivalent to dense model as in \textit{Baseline-1}. However, we  find that the performance of the dense network (\textit{this work}) is slightly below the performance of the dense (\textit{Baseline-1}) for CIFAR-10/STL-10/ImageNet-100 while for CIFAR-100 the performance is better than the self-supervised dense module.
This is interesting because what we achieve from this single pre-training of 500 epochs, is a single \underline{base encoder} (dense encoder) and gates (via gating modules) and their combination gives a gated \underline{lightweight} network.

\noindent \textbf{2. Transfer Learning:} Table~\ref{table2}  compares the transfer performance of VDG (dense and gated) with VICReg. This experiment gives further  insights into the quality of the learned representation in this joint setting. In general, there is a drop in performance for VICReg-Dual for both \textit{dense} and \textit{gated}, although the difference is not significant. However, for CIFAR-100~$\rightarrow$~CIFAR-10 \textit{dense} and \textit{gated} outperforms only \textit{dense} in VICReg at a very low target budget. Even in the case when the model is pre-trained on ImageNet-100, performance is comparable. This is encouraging as this new perspective still maintains good generalisation and transferability. 
\begin{figure*}[t]
    \centering
    \includegraphics[width=1.00\linewidth]{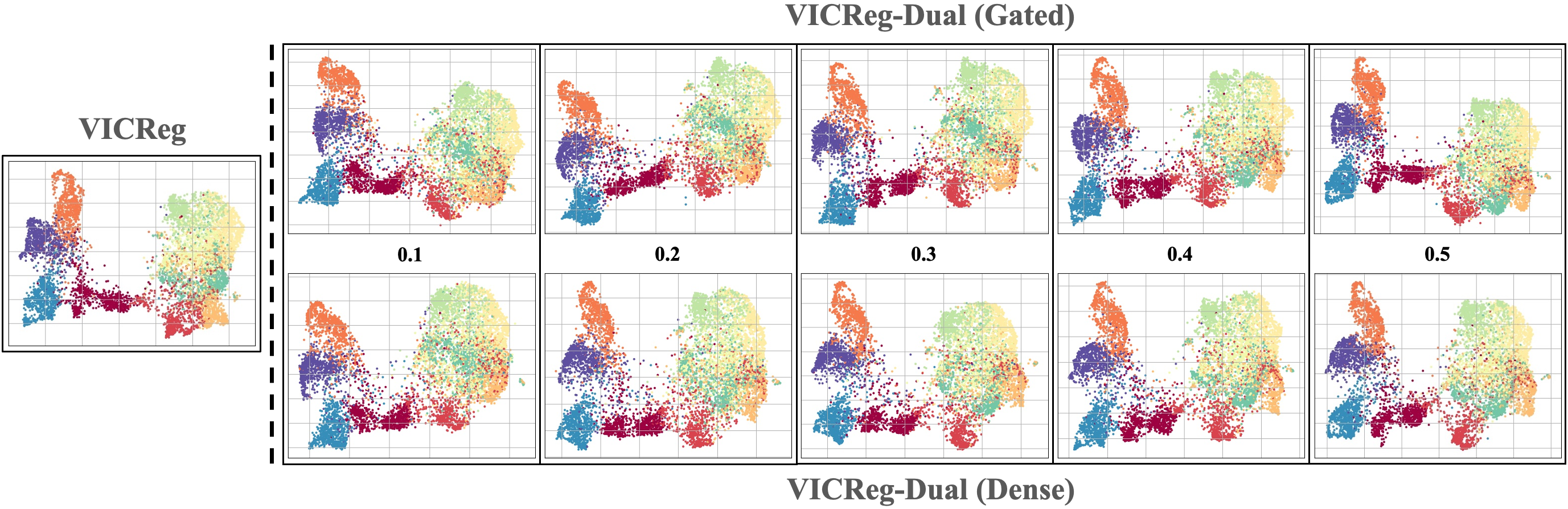}
    \vspace{-1.75\baselineskip}
    \caption{\textbf{Qualitative analysis:} UMAP embeddings of the learned representations: \textit{lightweight} gated network (\textit{top} row), while dense network (\textit{bottom}) row over different target budgets $t_{d}$. This is compared with embeddings of VICReg (dense) trained without any sort of sparsity. Best viewed in color. 
    }
    \label{fig:umap}
\end{figure*}

\noindent \textbf{3. SSL-KD vs SSL-Gating:} Table~\ref{tableKD} compares the performance of SSL-KD methods with our SSL-Gating framework. To avoid confusion, we would like to reiterate we don't follow student-teacher paradigm, ``Student'' in Table \ref{tableKD} for ``Ours (Gated)'' is basically a R18 (base-encoder) with gates (see Figure~\ref{fig:main}) and  $\mathcal{D}_{\text{Pretrain}} = \mathcal{D}_{\text{Target}}$. Results in Table~\ref{tableKD} are very promising as we outperform both the KD methods by a substantial margin across all budgets. This result suggests that combining gating could serve as a general recipe to obtain a lightweight network along with a \textit{dense} network during pre-training.

\noindent \textbf{4. Qualitative assessment:}
Figure \ref{fig:umap} shows uniform manifold approximation and projection (UMAP) \cite{mcinnes2018umap} embeddings of the learned representations ($f_{\theta}(\mathbf{x}) \in \mathbb{R}^{512}$) trained using the dual-setting on the STL-10 dataset and compares it with VICReg \cite{Bardes}. The learned structure is similar to dense (VICReg) at a very low budget. Furthermore, the classes appear to be visually distinct, similar to the VICReg setting, and this is observed for both the dense and gated networks of VICReg-Dual. 

\begin{wraptable}{L}{0.5\textwidth}
\vspace{-0.5cm}
 \caption{BT vs VICReg in dual setting.\Bstrut}
 \small
 \centering
 \resizebox{.5\textwidth}{!}{
 \begin{tabular}{l|c|c>{\columncolor[gray]{0.85}}cc|c>
 {\columncolor[gray]{0.85}}cc}
      
      \hline
      \hline
      \multicolumn{1}{l|}{\textbf{}} & \multicolumn{1}{l|}{} &
      \multicolumn{3}{c|}{\textbf{BT-Dual}}&
      \multicolumn{3}{c}{\textbf{VICReg-Dual}} \Tstrut\Bstrut\\
      
      \textbf{Dataset}&$\textbf{t}_{d}$ &\multicolumn{1}{c}{Dense} &\multicolumn{1}{c}{Gated} &FLOPs R. &\multicolumn{1}{c}{Dense} &\multicolumn{1}{c}{Gated} &FLOPs R. \Bstrut \\
\hline
&10\%&53.64 & 53.01 & 80.60\% &\textbf{54.92} & \underline{54.66} & \underline{81.71\%} \Tstrut\\
CIFAR-100&30\% & 55.03&54.89&66.09\%& \textbf{56.34} & \underline{55.53} & \underline{66.83\%}\\
&50\%& 55.63&55.39&47.64\% & \textbf{56.83} & \underline{57.25} & \underline{49.06\%}\\
\hline
&10\%&73.45& 73.49&76.74\%&\textbf{76.61}&\underline{76.45}& \underline{78.27\%} \Tstrut\\
STL-10&30\%&75.51&76.02&64.12\%&\textbf{78.55}& \underline{78.80}& \underline{65.17\%}  \\
&50\%&76.99&77.10&48.32\%&\textbf{79.69}&\underline{79.95}&\underline{48.39\%}\\
\hline
\hline
 \end{tabular}
 \label{table3}
}
 \vspace{-0.40cm}
 \end{wraptable}


\section{Additional Insights}
For all experimental settings and studies discussed hereafter; models (and settings) were pre-trained for 500 epochs on CIFAR-100 and STL-10 datasets. Instead of using linear evaluation we use kNN as the evaluation criterion with $k=1$ to report \textit{top-1} accuracy. In all tables, \textbf{bold} and \underline{underline} are the best performing results for the dense and gated module, respectively. Representations are not $l_{2}$ normalised.
\begin{wraptable}{R}{0.5\textwidth}
\vspace{-0.5cm}
 \caption{Alternative base encoders. \Bstrut}

 \small
 \centering
 \resizebox{.5\textwidth}{!}{
 \begin{tabular}{l|l|c>{\columncolor[gray]{0.85}}cc|c>{\columncolor[gray]{0.85}}cc}
      
      \hline
      \hline
      \multicolumn{1}{l|}{\textbf{}} & \multicolumn{1}{l|}{} &
      \multicolumn{3}{c|}{\textbf{VICReg-Dual}}&
      \multicolumn{3}{c}{\textbf{VICReg-Dual}} \Tstrut\Bstrut\\
      \multicolumn{1}{l|}{\textbf{}} & \multicolumn{1}{l|}{} &
      \multicolumn{3}{c|}{(1$\times$ResNet-18)}&
      \multicolumn{3}{c}{(2$\times$ResNet-18)} \Bstrut\\
      
      \textbf{Dataset}&$\textbf{t}_{d}$(\%) &\multicolumn{1}{c}{Dense} &\multicolumn{1}{c}{Gated} &FLOPs R. &\multicolumn{1}{c}{Dense} &\multicolumn{1}{c}{Gated} &FLOPs R. \Bstrut \\
\hline
&10\%& \textbf{54.92} & \underline{54.66} & 81.71\%&53.85 & 52.44 &\underline{85.77\%} \Tstrut\\
CIFAR-100&30\% & \textbf{56.34} & \underline{55.53} & 66.83\% & 55.65&55.52&\underline{70.07\%}\\
&50\%& \textbf{56.83} & \underline{57.25} & 49.06\%&55.64 &55.71 &\underline{52.60\%} \Bstrut\\
\hline
&10\%&\textbf{76.61}&\underline{76.45}& 78.27\%&74.93 &74.89&\underline{82.48\%} \Tstrut\\
STL-10&30\%&\textbf{78.55}& \underline{78.80}&65.17\%&76.83&76.71&\underline{69.26\%}  \\
&50\%&\textbf{79.69}&\underline{79.95}&48.39\%&77.74&78.00 &\underline{51.10\%}\\
\hline
\hline

 \end{tabular}
 \label{table4}
 }
 \vspace{-0.45cm}
 \end{wraptable}
  
\noindent \textbf{1.~Comparison with the symmetric Barlow Twins (BT) architecture.} VICReg is build upon the findings of BT \cite{Zbontar2021} and it is straightforward to apply the dual setting to BT because it minimises the cross-correlation (regularisation term) to identity $\mathcal{I}$ although the loss function is entirely mutual unlike in VICReg. 
In Table~\ref{table3} we compare the performance of BT augmented with \textit{our} setting. We observe that 1-NN performance of BT is low as compared to VICReg-Dual. The drop in performance could be attributed to the fact that VICReg applies independent regularisation which are later matched through the invariance loss. This further validates the hypothesis of choosing VICReg as our objective.

\begin{wraptable}{L}{0.5\textwidth}
\vspace{-0.450cm}
 \caption{Investigating the role of MSE. \Bstrut}

 \small
 \centering
 \resizebox{.5\textwidth}{!}{
 \begin{tabular}{l|l|c>{\columncolor[gray]{0.85}}cc|c>{\columncolor[gray]{0.85}}cc}
      
      \hline
      \hline
      \multicolumn{1}{l|}{\textbf{}} & \multicolumn{1}{l|}{} &
      \multicolumn{3}{c|}{\textbf{VICReg-Dual}}&
      \multicolumn{3}{c}{\textbf{VICReg-Dual}} \Tstrut\Bstrut\\

      \multicolumn{1}{l|}{\textbf{}} & \multicolumn{1}{l|}{} &
      \multicolumn{3}{c|}{w/ Invariance}&
      \multicolumn{3}{c}{w/o Invariance} \Bstrut\\
      
      \textbf{Dataset}&$\textbf{t}_{d}$(\%) &\multicolumn{1}{c}{Dense} &\multicolumn{1}{c}{Gated} &FLOPs R. &\multicolumn{1}{c}{Dense} &\multicolumn{1}{c}{Gated} &FLOPs R. \Bstrut \\
\hline
&10\%& \textbf{54.92} & \underline{54.66} & 81.71\%&4.06 & 6.31 &\underline{93.77\%} \Tstrut\\
CIFAR-100&30\% & \textbf{56.34} & \underline{55.53} & 66.83\% & 4.84&4.89&\underline{82.72\%}\\
&50\%& \textbf{56.83} & \underline{57.25} & 49.06\%&2.79 &4.15 &\underline{49.05\%} \Bstrut\\
\hline
&10\%&\textbf{76.61}&\underline{76.45}& 78.27\%&11.79 &18.02&\underline{90.71\%} \Tstrut\\
STL-10&30\%&\textbf{78.55}&\underline{78.80}&65.17\%&12.25&17.89&\underline{74.96\%}  \\
&50\%&\textbf{79.69}&\underline{79.95}&48.39\%&12.81&15.72&\underline{51.91\%}\\
\hline
\hline

 \end{tabular}
 \label{co-evolve}
 }
 \vspace{-0.450cm}
 \end{wraptable}
\noindent \textbf{2.~Training with a different base encoder.} In this setting we train a model with two base encoders, one w/ gate (gated) and other w/o gate (dense) i.e, ($2 \times \text{ResNet-18}$) (results in Table~\ref{table4}). An interesting observation is that VICReg-Dual with a single base encoder outperforms a more powerful setting with two base encoders (Table~\ref{table4}) although the FLOPs reduction (FLOPs R.) is higher ($\uparrow$) in the setting of two different encoders. This is due to the fact that the sparsity loss ($\mathcal{L}_{G}$) operates solely on the un-shared branch so there is no trade-off involved, as in case of single base encoder which simultaneously tries to enforce sparsity and visual invariance.

\noindent \textbf{3.~Role of \textit{mean squared error} in co-evolving.} It's a well known fact in SSL that these methods suffer from dimensional collapse \cite{hua2021feature, Jing}. Training without any regularisation term or any trick \cite{grill2020bootstrap, chen2021exploring} would lead to dimensional collapse. Also, the authors in \cite{kim2021comparing} showed that MSE serves as a better option for exact logit matching as compared to Kullback-Leibler (KL) divergence. So, in order to understand the role of MSE we trained a model w/o the \textit{invariance} loss. As Table \ref{co-evolve} shows, we found that there is large performance drop if we remove the invariance term. This implies that the invariance term plays a crucial role and seem to be an important factor, not only for co-evolving, but for self-supervision.

 \section{Discussion and Conclusion}
In this work we presented a novel perspective on unifying synergies between SSL and DC. We exploit DC to induce sparsity into symmetric branches of self-supervised models enabling both  branches to co-evolve with each other during training. In addition, this approach also allows simultaneous training of a dense and gated (sparse) sub-network from scratch with a target budget $t_d$ under a self-supervised training objective with minimal computational overhead via weight sharing, thereby offering a good accuracy-efficiency trade-off for a given downstream application. As a result, our single base encoder offers enough flexibility to serve a dual purpose to reduce excessive computational overhead, which we validated through exhaustive experimentation (Tables~(\ref{table1}, \ref{table2}, \ref{tableKD})).
However, there are limitations with this work. First, the dense model performance degrades and its performance further fluctuates with varying $t_{d}$. We experimented with RotNet \cite{gidaris2018unsupervised} as a extra proxy loss for the dense branch but it did not yield good performance. 
Second, we have not imposed any constraint in the training objective that enforces a uniform distribution of channel activations, i.e. preservation of channel diversity during inference (which could also be a solution to the first limitation). Third, in future we will extend this setting to contrastive and non-symmetric architectures. \\
The work in this paper is an initial attempt to  draw parallels and make an inter-connection between both of these fields. However, more research is needed to build a better intuition and insight into these models and also help us understand their other attributes, such as generalisation and ability to transfer to other downstream tasks. 

\section{Acknowledgments}
This publication has emanated from research conducted with the financial support of Science Foundation Ireland under Grant number SFI/12/RC/2289\_P2, co-funded by the European Regional Development Fund and Xperi FotoNation. This work is dedicated to the memory of Dr.~Kevin McGuinness, whose contributions will always be remembered. This work stands as a testament to his enduring influence and our deep appreciation for his invaluable contributions.

%

\bibliography{sample}
\end{document}


\maketitle




\section{Data Augmentation}
We follow the augmentation strategy as outlined in VICReg \cite{Bardes} which remains a standard augmentation strategy in almost every SSL scenario. The following augmentations were applied sequentially: 
\begin{itemize}
    \item Random cropping with an area uniformly sampled with size ratio between 0.2 to 1.0, followed by resizing to size: $(224 \times 224)$ (ImageNet-100), $(96 \times 96)$ (STL-10) and $(32 \times 32)$ (CIFAR-10/100).
    \item Random horizontal flip with $p=0.5$\footnote{$p$ denotes probability.}.
    \item Color jittering of brightness, contrast, saturation and hue, with  $p=0.8$, with 
ColorJitter params as (0.4, 0.4, 0.2, 0.1).
\item Grayscale with $p=0.2$.
\item Solarization with $p=0.1$.
\end{itemize}

\section{Learned Policy: Qualitative assessment}
\begin{figure*}
   \centering
  \includegraphics[width=1.0\textwidth]{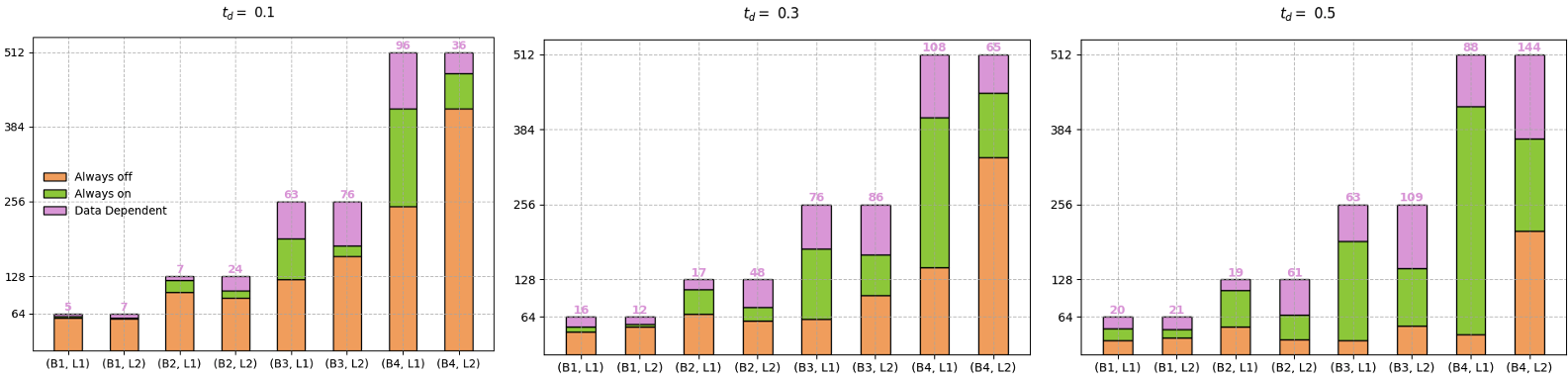}  
  \caption{Learned channel distribution for CIFAR-100 with varying $t_{d}$. }
  \label{fig:cifar} 
\end{figure*}

\begin{figure*}
   \centering
  \includegraphics[width=1.0\textwidth]{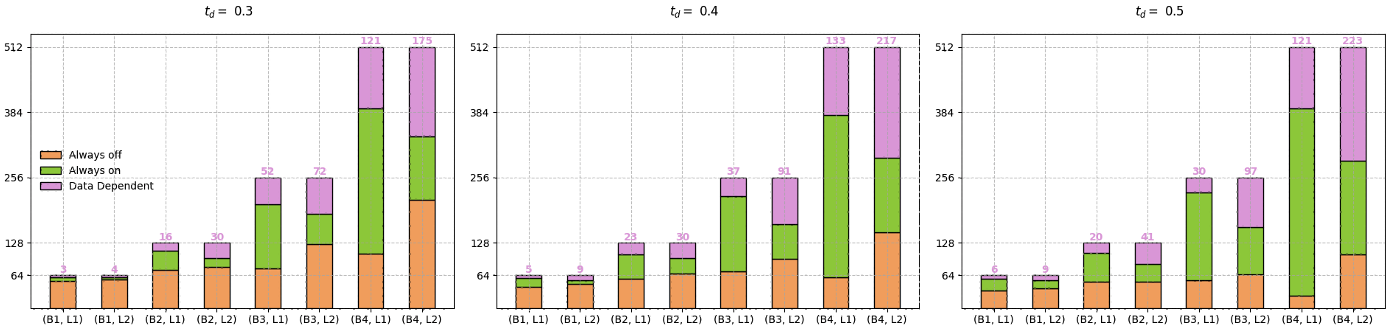}  
  \caption{Learned channel distribution for ImageNet-100 with varying $t_{d}$}
  \label{fig:img} 
\end{figure*}

Figure \ref{fig:cifar} and \ref{fig:img} depicts the learned channel distribution through our gating module. In order to make fair comparison with Knowledge distillation method specifically SEED \cite{fang2021seed} and SimReg \cite{navaneet2021simreg}, student networks were sampled from R18's subspace following policy learned by our gated network Figure \ref{fig:cifar} \& \ref{fig:img}. We combine ``data dependent'' + ``always on'' channels to get the architecture of student for uniform comparison. 


\bibliography{sample}